\documentclass[letterpaper]{article}
\usepackage{aaai20}
\usepackage{times}
\usepackage{helvet}
\usepackage{courier}

\usepackage[dvipsnames]{xcolor}
\definecolor{mypink1}{rgb}{0.858, 0.188, 0.478}

\usepackage{color}
\definecolor{blue}{RGB}{0, 93, 170}

\usepackage{graphicx}
\usepackage{arydshln}
\usepackage{booktabs}
\usepackage{todonotes}

\usepackage{soul}
\usepackage{url}
\usepackage[hidelinks]{hyperref}
\usepackage[utf8]{inputenc}
\usepackage[small]{caption}
\usepackage{amsmath}
\usepackage{amsfonts}
\usepackage{placeins}

\usepackage{color}
\usepackage{todonotes}
\definecolor{blue}{RGB}{0, 93, 170}			%Go Big Blue!
\definecolor{darkgreen}{RGB}{0, 102, 0}

\newcommand{\eat}[1]{}
\newcommand{\defn}[1]           {{\textit{\textbf{\boldmath #1}}}}
\newcommand{\topic}[1]       {{\smallskip\noindent{\textbf{\boldmath #1 \ }}}}

\begin{document}

\title{Generalizable Resource Allocation in Stream Processing\\ via Deep Reinforcement Learning}

\author{
Xiang Ni$^\clubsuit$\thanks{This work was done when the author was at IBM.}\quad
Jing Li$^\heartsuit$\quad
Mo Yu$^{\spadesuit}$\quad
Wang Zhou$^{\spadesuit}$\quad
Kun-Lung Wu$^{\spadesuit}$\\
% \affiliations
$^\clubsuit$Citadel\quad $^\spadesuit$IBM Research\quad $^\heartsuit$New Jersey Institute of Technology\\
xiang.ni@citadel.com,
jingli@njit.edu,
yum@us.ibm.com,
wang.zhou@ibm.com,
klwu@us.ibm.com
}

\maketitle

\thispagestyle{plain}
\pagestyle{plain}
\renewcommand{\thepage}{\arabic{page}}
\setcounter{page}{1}

\begin{abstract}
This paper considers the problem of resource allocation in stream processing, where continuous data flows must be processed in real time in a large distributed system. To maximize system throughput, the resource allocation strategy that partitions the computation tasks of a stream processing graph onto computing devices must simultaneously balance workload distribution and minimize communication. Since this problem of graph partitioning is known to be NP-complete yet crucial to practical streaming systems, many heuristic-based algorithms have been developed to find reasonably good solutions. In this paper, we present a graph-aware encoder-decoder framework to learn a \textit{generalizable} resource allocation strategy that can properly distribute computation tasks of stream processing graphs unobserved from training data.
We, for the first time, propose to leverage graph embedding to learn the structural information of the stream processing graphs. Jointly trained with the graph-aware decoder using deep reinforcement learning, our approach can effectively find optimized solutions for unseen graphs. Our experiments show that the proposed model outperforms both METIS, a state-of-the-art graph partitioning algorithm, and an LSTM-based encoder-decoder model, in about $70\%$ of the test cases.

\end{abstract}

\section{Introduction}

\begin{figure}[!tbp]
\centering
% \vspace{-4mm}
\includegraphics[width=1.0\columnwidth]{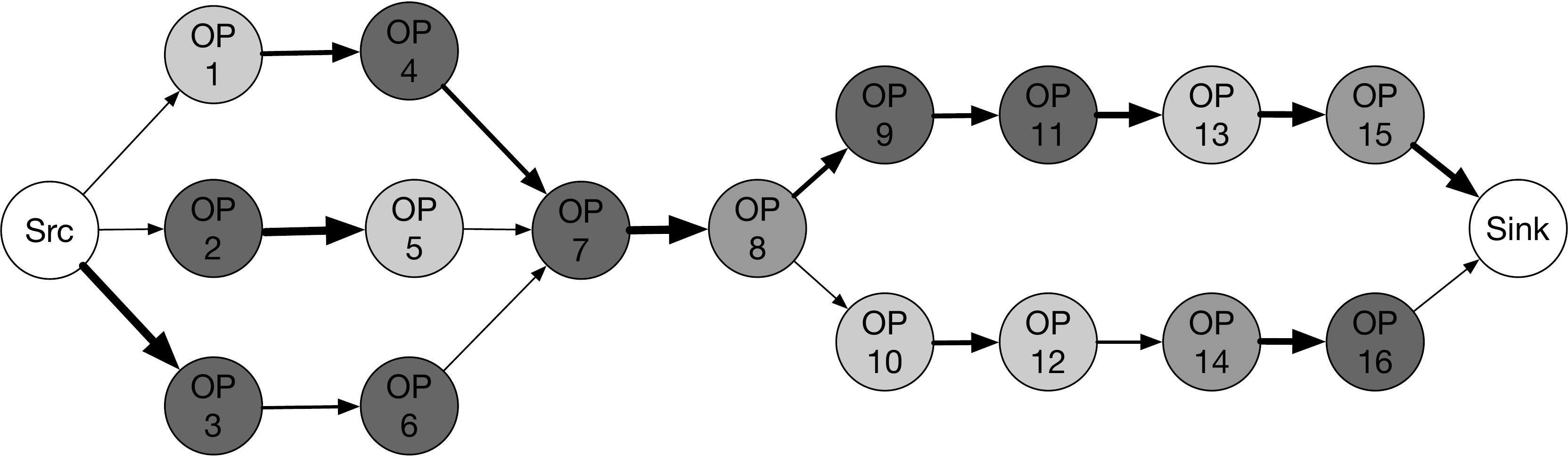}
\vspace{-6mm}
\caption{An example stream processing graph. The nodes of the graph represent operators labeled with the amount of computation work. The darker color of an operator illustrates higher CPU utilization. The directed edges depict data flowing from the source to destination operators. The edge width represents the amount of data flowing through that connection.} 
\vspace{-2mm}
\label{fig:sample-graph}   
\vspace{-0.1in}
\end{figure}

In various industrial domains, such as aviation, medicine, transportation, telecommunication, and banking, online stream processing has been widely used to analyze high-rate data with high throughput and generate live results in a timely manner~\cite{streams,flink}.
The computation in stream processing is driven by incoming data flowing through a Directed Acyclic Graph (DAG)\footnote{Some systems, e.g., \cite{streams}, allow cycles.}. A \defn{stream graph} is comprised of \defn{operators}, which conduct computation on the incoming tuples, and directed \defn{edges}, each of which connects two operators and transmits tuples between them. \defn{Tuples} are structured data items with strongly-typed attributes. Operators are event-driven and execute only when there is a tuple received. Figure~\ref{fig:sample-graph} shows an example of a stream graph.

To exploit the abundant parallelism in stream graphs, operators are distributed to computing devices (resources, e.g., CPUs or GPUs) for execution. One objective in a stream processing system is to maximize the \defn{throughput} --- the number of tuples processed per second. If two connected operators are distributed to the same device, they can communicate via a simple function call \cite{streams}. If they are distributed to different devices, however, they need to communicate over a network. As function calls are more efficient than network communications, it is important to properly distribute operators to computing devices. A good resource allocation strategy should well balance the trade-offs between distributing computation evenly across devices and minimizing communication cost between devices.

The distribution of operators to resources can be formulated as a two-step \defn{graph partitioning} problem: first determining the right number of partitions $k$ and then performing $k$-way partitioning to divide operators into $k$ sets\footnote{This is referred to as the \emph{fusion} of operators into Processing Elements (PEs) in \cite{streams}, where each set is a PE.}. Finding the optimal graph partition is known to be NP-complete. In practice, only approximate solutions are possible with the adoption of heuristic rules~\cite{karypis1998fast}. Generally, existing graph partitioning libraries share a common drawback: they usually fail to give a good approximation of the minimum number of partitions $k$ and may not consistently generate high-quality partitions, since $k$ is usually graph-dependent but the structure of graphs can change from application to application. Analytical performance modeling and prediction have also been proposed~\cite{li2016performance}. However, existing models either have strong assumptions of data arrival rate and node connectivity, or fail to fully capture the complex factors affecting the data processing throughput in practical stream processing systems. 

This paper focuses on learning a \defn{generalizable resource allocation strategy} that can intelligently partition stream graphs with \textit{different} structures while capturing performance-relevant factors of practical stream processing systems. Our proposed graph-aware encoder-decoder framework is able to produce optimized solutions to graphs that are not observed in the training data. In particular, we make the following contributions.

\topic{DRL as a solution for graph partitioning.}
We conduct the first study of using deep reinforcement learning (DRL) to train a good graph partitioning strategy that is generalizable to different stream graphs. Since graph partitioning is NP-complete, a general approximation approach is to reformulate it as a search problem for partitioned graphs. DRL has been proven to be a good mechanism for improving the search strategy in structured prediction problems, e.g., natural language syntactic parsing~\cite{chang2015learning} and neural architecture search for discovering better neural network architectures than human-designed ones~\cite{zoph2016neural,liu2017progressive}.
Inspired by these successes, we apply DRL to graph partitioning on stream processing systems. 
Our DRL solution is formulated as a sequence of resource allocation predictions. Each step of the prediction makes use of the state representations on relations between the current assignment and the global property of the stream graph, in order to approximate the global search rewards.
Therefore, our approach can benefit from the capability of deep networks to efficiently learn the state representations of stream graphs and partial allocations, as well as their relations in a search policy.

\topic{Generalizable model with graph-aware state representations.}
A good graph partitioning strategy should be generalizable to different input graph structures. Therefore, to learn a search policy for unseen graphs, the DRL solution must be able to capture the desired properties of graph topology into the state representations and the state representation space must be transferable among different graphs. In stream graphs, such properties include the global information of the entire graph, such as each operator's position in the graph, the critical-path of the graph, the relation between operators' CPU utilization and communication cost, and the relation between devices' assignment predictions.
To this end, we make the state representations be aware of (1) the graph topology, by leveraging the graph embedding generated by the recent advance in graph neural networks~\cite{kipf2016semi,hamilton2017graphsage,xu2018graph2seq}; and (2) the predicted assignment history, by enhancing the prediction of each device with its relevant graph neighbors' predicted assignments.

\topic{New benchmark and evaluation.}
To evaluate the proposed approach, we construct a benchmark\footnote{Our code and data set are released at \url{https://github.com/xiangni/DREAM}.} containing $3,150$ stream graphs that are representative of different computation and communication requirements. In addition, we conduct the evaluation in a realistic setting where the graphs in testing are unobserved from training.
We compare our proposed graph-based DRL method against two well-regarded solutions: METIS~\cite{karypis1998fast}, a state-of-the-art graph partitioning library, and an LSTM-based encoder-decoder model designed to optimize the device placement for one fixed graph~\cite{mirhoseini2017device}.
Our experiments show that our proposed approach achieves superior performance by improving resource allocations for $76\%$ of graphs and increasing average throughput by $12\%$.

\section{Problem Definition}
\label{sec:background}
\begin{figure}[!h]
\vspace{-2mm}
\includegraphics[trim=0cm 1cm 0cm 0cm, width=\columnwidth]{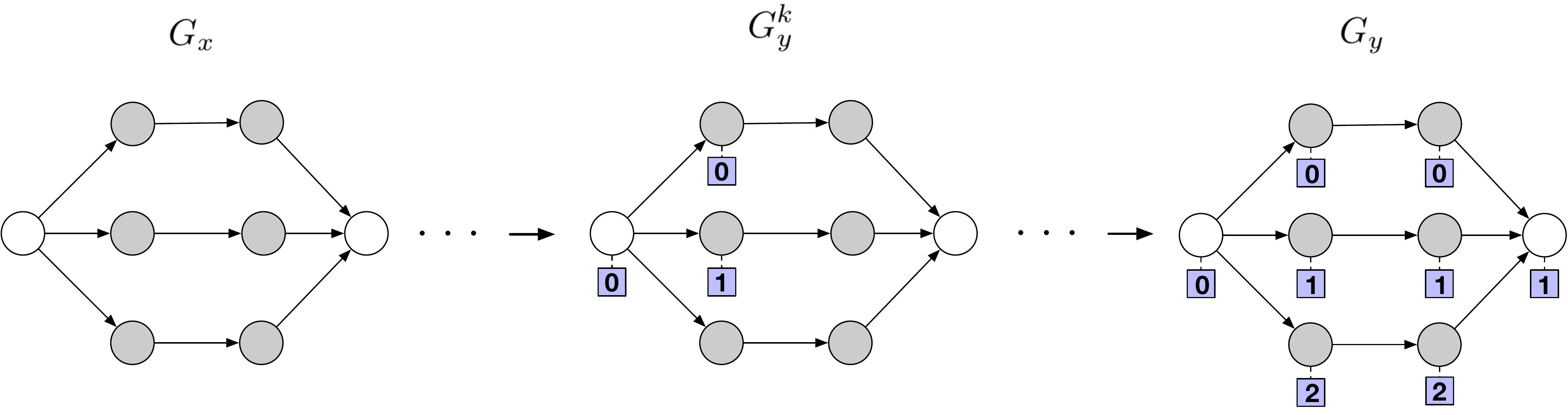}
\caption{Graph-to-graph generation process. Square nodes indicate that the connected operators have been allocated to that device.}
\label{fig:g_x}   
\vspace{-2mm}
\end{figure}

This section formally defines the resource allocation problem in stream processing and formulates it as a search problem.
Our training and testing data consists of different stream graphs $\{G_x\}$.
The computation of a stream processing task forms a graph $G_x = (V,E)$, as illustrated in 
the left graph of Figure \ref{fig:g_x}. Each node $v \in V$ is an operator characterized by its \defn{CPU utilization} (number of instructions required per second) and \defn{payload} (total size of tuples produced by the operator). Each directed edge $e = (v_{in}, v_{out}) \in E$ represents the connection between operators $v_{in}$ and $v_{out}$, via which $v_{in}$ transmits its output tuples to $v_{out}$ as input. Each edge is labeled with its communication cost.

The goal of the resource allocation given an input graph $G_x$ and a set of devices (e.g., CPUs) $D$ is to predict a \defn{device placement graph} $G_y$ where each operator $v\in G_x$ is assigned to a device $d_v \in D.$ 
We aim to train a “meta” model on a set of stream graphs that is able to make good resource allocation predictions for \defn{unseen stream graphs}: graphs with different topology or different distribution of operator CPU utilization and payload compared to graphs in the training data set.
In this work, we focus on homogeneous devices where the device ids can be interchangeable and leave the case of heterogeneous devices as future work.
The right graph in Figure \ref{fig:g_x} illustrates the target graph $G_y$ where each node in $G_x$ is appended with a new \defn{device id node} (the square nodes in the figure) depicting its allocation.

Our task of predicting $G_y$ is challenging because of the intertwined dependencies between device allocations. 
Intuitively, the placement $d_{v_i}$ of a node ${v_i}$ depends not only on \emph{the topology of the input graph $G_x$}, but also on \emph{the placements of the other nodes}.
Formally, the prediction follows $P(d_{v_i}|G_x, d_{v_{i-1}}, d_{v_{i-2}}, d_{v_{1}})$, which is difficult to be decomposed.
Hence, our task requires joint inference of the whole graph $G_y$. We formulate this joint inference problem as a search problem that appends one device id node to each operator node in $G_x$ at each step, and finally outputs the graph $G_y$.
Because of the high dependencies among the previously predicted $d_v$s, our task is much harder compared to classification over single nodes, and it is closer to a special case of graph-to-graph generation, where the output $G_y$ contains one additional device id node for each operator node in $G_x$.

Moreover, the difficulty of our setting also lies in the diverse input graphs with complex topologies in training and testing.
The diversity makes it hard for a model to memorize the graph topologies like previous work where training and testing are done on one single graph~\cite{mirhoseini2017device}. Thus, a generalizable way to represent the many different graphs is necessary.
In this work, we design and train our model to capture the meta-information of graph topologies so that it can directly predict good placements for various graphs, without requiring to individually train a different model for each of the different graphs as needed in~\cite{mirhoseini2017device}.

\begin{figure*}[!htbp]
\centering               
\includegraphics[width=2\columnwidth]{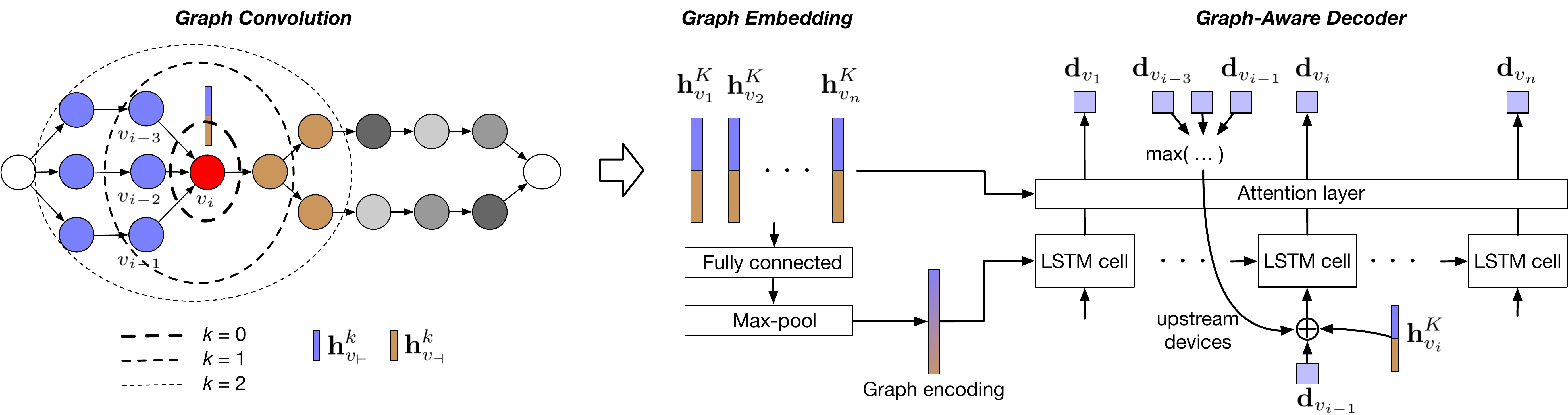} 
\vspace{-2mm}
\caption{Overview of the proposed model architecture. Our model is composed of graph encoding to learn the structured information of stream graphs and graph-aware decoder for device allocation prediction.}
\vspace{-4mm}
\label{fig:model_overview}  
\end{figure*}

\section{Graph-Aware Encoder-Decoder Model}
\label{sec:model}

Figure \ref{fig:model_overview} shows the overall architecture of our proposed approach.
Our model first encodes the input stream graph with a graph encoder (Section \ref{ssec:encoder}). The resulted graph embedding is then used for a graph-aware decoder to generate device assignments in a sequential way (Section \ref{ssec:decoder}). We optimize the entire network with DRL (Section~\ref{ssec:training}).

\subsection{Stream Graph Encoding}
\label{ssec:encoder}

Our model first embeds the input graph $G_x$ to an embedding space. Each node has an embedding encoding its contextual information in the graph that is informative to its placement prediction.
We achieve such encoding using the Graph Convolution Network (GCN)~\cite{kipf2016semi}.

GCN iteratively updates a node's embedding (hidden states) with its neighbors' embeddings.
Specifically, at the $k$th step, for each operator node $v$, we define its embedding as $\textbf{h}^k_v$. When $k$=0, $\textbf{h}^k_v$ is defined as its node feature vector $\textbf{f}_v$, which contains the CPU utilization and payload of the tuples emitted from this node. 
Because $G_x$ is a directed graph, according to the directions of $v$'s edge connections, we categorize its neighbors into two sets, the upstream neighbors $\mathcal{N}_{\vdash}(v)$ and downstream neighbors $\mathcal{N}_{\dashv}(v)$. The node embedding thus can be categorized as two vectors $\mathbf{h}^k_{v_{\vdash}}$ and $\mathbf{h}^k_{v_{\dashv}}$, each with $m$ dimensions. 

Based on the above definitions, following \cite{hamilton2017graphsage,xu2018graph2seq}, the GCN updates $v$'s embedding as below:

\noindent$\bullet$ First, we aggregate the information from $v$'s upstream and downstream neighbors separately. Taking the aggregation of upstream neighbors for example, for each $u \in \mathcal{N}_{\vdash}(v)$, we take its current stage representation $\mathbf{h}^k_{u\vdash}$, feed it to a non-linear transformation $\mathbf{h}^{\mathrm{(in)}}_{u\vdash} = \mathrm{tanh} (\mathbf{W^{(up)}_1} \mathbf{h}^k_{u\vdash})$,\footnote{Ideally, we have $\mathbf{h}^{\mathrm{(in)}}_{u\vdash} = \mathrm{tanh} (\mathbf{W^{(up)}_1} [\mathbf{h}^k_{u\vdash}$:$\mathbf{f}_e(u,v)])$, where we derive edge features $\mathbf{f}_e(u,v)]$ for prediction. However, empirically we have payload features on nodes that correlates with edge weights. Thus concatenating the edge weight features does not help. We leave investigating richer edge features to future work.} where $\mathbf{W^{(up)}_1} \in \mathbb{R}^{m \times 2m}$.

\noindent$\bullet$ Second, we get all the $\mathbf{h}^{\mathrm{(in)}}_u, \forall u \in \mathcal{N}_{\vdash}(v)$, take the mean-pooling of the vectors and update the upstream-view embedding of $v$ as ($[\cdot:\cdot]$ refers to vector concatenation):
\begin{align}
\mathbf{h}^{k+1}_{v_{\vdash}} = \mathrm{tanh} (\mathbf{W^{(up)}_2} \left[\mathbf{h}^{k}_{v_{\vdash}}:\dfrac{\sum_{u\in \mathcal{N}_{\vdash}(v)}\mathbf{h}^{\mathrm{(in)}}_u}{\vert\mathcal{N}_{\vdash}(v) \vert}\right]).
\end{align}

Similar update is applied to the \textbf{downstream} representation of $v$ to get \textbf{h}$_{v\dashv}^{k+1}$, which operates on the $\mathcal{N}_{\dashv}(v)$ neighbors with %separated 
transformation parameters $\mathbf{W^{(down)}_1}$ and $\mathbf{W^{(down)}_2}$.
In our experiments, we use shared parameters for upstream and downstream updates.

The above steps are repeated $K$ times over all nodes in the graph. Finally for each $v$ we concatenate its upstream and downstream hidden states $\mathbf{h}^K_{v_{\vdash}}$ and $\mathbf{h}^K_{v_{\dashv}}$ as its final node representation. We denote the vector as $\mathbf{h}_{v}$ for short in the following sections. 

We further compute the graph encoding to convey the entire graph information. The embedding of each node $\mathbf{h}_{v}$ is fed to a fully connected neural network layer, followed by an element-wise max-pooling layer. The output vector is thus graph encoding, used as input to the graph-aware decoder, as illustrated in Figure~\ref{fig:model_overview}.

\subsection{Graph-Aware Decoding of Device Allocation}
\label{ssec:decoder}

The prediction of the resource allocation is to assign each operator node $v$ in $G_x$ to a device $d_v$, conditioning on the graph property and the assignments of other nodes. 
Given an arbitrary order of nodes $\{v_1, v_2, ..., v_{\vert V\vert}\}$ from $G_x$,
the problem can be formulated as:
\begin{align}
\label{eq:decoding}
    P(G_y=\{d_{v_1}, d_{v_2}, \cdots, d_{v_{\vert V \vert}}\}|G_x)\\ \nonumber
    =\prod_{t}P(d_{v_t}|d_{v_{t-1}}, d_{v_{t-2}}, \cdots, d_{v_{1}}, G_x). \nonumber
\end{align}
The joint probability cannot be trivially decomposed, as the dependency between the new assignment $d_{v_t}$ and all previous assignments highly depend on the property of graph $G_x$.

In the decoding stage, we adopt an approximated decomposition of Eq.~(\ref{eq:decoding}) in order to simplify the problem.
Intuitively, the device prediction of one node is usually highly influenced by the device assignments of its upstream nodes.
Therefore, if we could always have a node $d_v$'s upstream nodes assigned before it (e.g. ordering the nodes via breadth-first traversing of the graph), we can have the following approximation:
{
\begin{align}
    P(G_y|G_x)    =\prod_{t}P(d_{v_t}|D^{(up)}(v_t), G_x),
    \label{eq:decoder_decomp}
\end{align}
}%
where $D^{(up)}(v_t)$ refers to the assignments of all the upstream nodes of $v_t$.
Our proposed graph-aware decoder is based on this decomposition.

\topic{States Representation in Decoder.}
To deal with the intertwined dependencies among the new assignment $d_{v_t}$, all previous assignments and the $G_x$, the DRL model learns a state representation $\mathbf{s}_t$ to encode the information associated with $\{d_{v_{t-1}}, d_{v_{t-2}}, \cdots, d_{v_{1}}, G_x\}$. 
This can be implemented with an LSTM~\cite{hochreiter1997long}: 
\begin{align}
    \mathbf{s}_{t} = \mathrm{LSTM\_Cell}(\mathbf{h}_{v_t}, \mathbf{s}_{t-1}, d_{v_{t-1}}). \nonumber
\end{align}

However, although LSTM is proposed to memorize long-term temporal dependency, in practice it is difficult to learn an LSTM to memorize the history well without further inductive biases.
To further encourage the state $\mathbf{s}_t$ be aware of the assignments of the nodes related to $v_t$ to help its assignment prediction, 
inspired by the decomposition in Eq.~(\ref{eq:decoder_decomp}), we model the state of our decoder as:
\begin{align}
    \mathbf{s}_{t} = \mathrm{LSTM\_Cell}(\mathbf{h}_{v_t}, D^{(up)}(v_t), \mathbf{s}_{t-1}, d_{v_{t-1}}).\nonumber
\end{align}

In our implementation, we convert all the device assignments $d_v$s to their (learnable) device embedding vectors. The set $D^{(up)}(v_t)$ is thus represented as the mean-pooling results of the device embeddings to concatenated to the LSTM inputs.
The above vectors are concatenated with the device embedding of $d_{v_{t-1}}$ as input to the LSTM cell.

\topic{Prediction Layer.}
Finally, the prediction for node $v_i$ becomes $P(d_{v_i}|\mathbf{s}_i)$.
We model this step with an attention-based model \cite{bahdanau2014neural} over all the graph nodes $v_j$s. Each node receives an attention score at step $t$ as $\alpha_j = \mathbf{s}_t^T \mathbf{h}_{v_j}^K$. Then all $\alpha_j$s get normalized with softmax, leading to $\tilde{\alpha_j}$ for each $j$.
Finally, to make the device prediction, we feed the concatenation $[\mathbf{s}_t:\sum_j \tilde{\alpha_j} \mathbf{h}_{v_j}^K]$ to a multi-layer perceptron (MLP) followed by a softmax layer.

\subsection{Training}
\label{ssec:training}

In our task, it is difficult to get the ground truth allocation $G_y$ for an input $G_x$. However, given any allocation $G_y$, we can obtain its relative quality by calculating the throughput. Therefore, our task fits the reinforcement learning setting, where the model makes a sequence of decisions (i.e., our decoder) and gets delayed reward $r$ (i,e., the throughput of the predicted graph allocation).

In this work, we seek to maximize the \defn{relative throughput} $r(G_y) = \frac{T(G_y)}{I(G_x)}$, which is defined as the ratio of throughput $T(G_y)$ to the source tuple rate $I(G_x)$. This is because the objective is to ensure the tuple processing rate (throughput) catches up with the source tuple rate, i.e., no backpressure due to bad resource allocation. Hence, the range of reward $r$ is between $0$ and $1$. We train a stochastic policy to maximize the following objective, in which $\pi_\theta$ is a distribution over all possible resource allocation schemes $Y$:
\begin{align}
J(\theta) = \sum_{G_y \in Y}\pi_\theta(G_y)r(G_y)
\nonumber
\end{align}

We apply the REINFORCE algorithm~\cite{williams1992simple} to compute the policy gradients and learn the network parameters using Adam optimizer~\cite{kingma2014adam}
\vspace{-2mm}
\begin{align}
\nabla(\theta)J(\theta) = \frac{1}{N}\sum_{n=1}^{N}\nabla\log\pi_\theta(G^{n}_{y})\big[r(G^{n}_{y})-b\big]\label{eq:reinforce}
\end{align}
\vspace{-2mm}

In each training update, we draw a fixed number of on-policy samples. We also draw random samples to explore the search space, and the number of random samples exponentially decays. 
Due to the sparsity of good resource allocation schemes over the large search space, for each training graph, we maintain a memory buffer~\cite{liang2018memory} to store the good samples with reward higher than $0.8$.
Extra random samples are included to fasten the exploration if the memory buffer is empty.
$N$ samples in Eq.~\ref{eq:reinforce} are composed of both on-policy samples and those from the memory buffer. Baseline $b$, the average reward of the $N$ samples, is subtracted from the reward to reduce the variance of policy gradients.  

\topic{Remark} 
Our proposed method learns from training data that are representative of real streaming graphs in terms of similar topologies and distributions but not the same graphs as in testing data. 
If the topologies in testing are not commonly seen in streaming workloads and hence not sufficiently covered by the training data, our model could be further trained. For example, upon deployment in operation, we could perform periodic incremental training with additional graphs seen in history. We leave this study to future work.

\begin{figure*}[!tb]
\centering
\includegraphics[trim=0cm 0.7cm 0cm 0cm, width=0.8\textwidth]{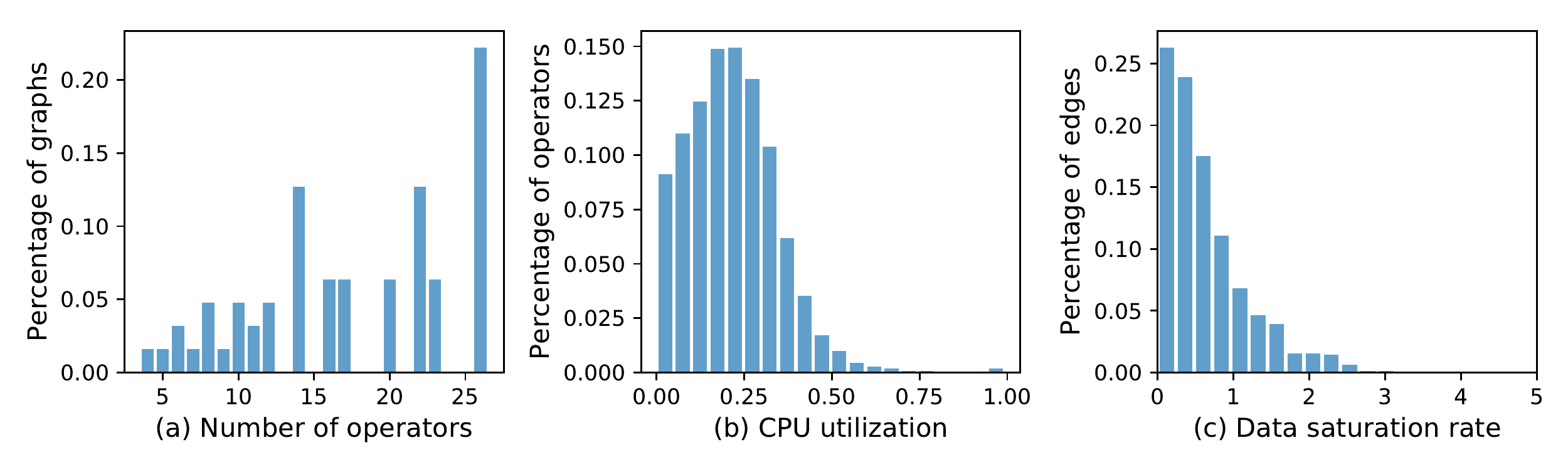}
\caption{Statistics of graphs in our benchmark data set.} 
\label{fig:graph_stats}  
\vspace{-4mm}
\end{figure*}

\subsection{Speeding up the Reward Calculation}
To compute the reward $r(G_y)$, each sampled allocation $G_y$ needs to be deployed on the stream processing system, however, the process may take up to a few minutes for the system to stabilize and calculate the throughput.
The total time and computing resource required for training in this way is simply intractable, given that DRL relies on the evaluation of numerous resource allocation trials.

Therefore, for fast training, we adopt a simulator \defn{CEPSim} for complex event processing and stream processing~\cite{higashino2016cepsim} to evaluate each allocation sample.   
CEPSim is a simulator for cloud-based complex event processing and stream processing system that can be used to study the effect of different resource allocation, operator scheduling, and load balancing schemes. 
In CEPSim, DAGs are used to represent how input event streams are processed to get complex events. CEPSim provides the flexibility for users to specify the number of instructions per tuple for each operator in the DAG. To simulate stream processing, users can give a mapping function to allocate parts of the DAG to different virtual machines (VMs), and these VMs can communicate with each other using a network. We extend CEPSim to allow users to specify the payload of tuples emitted by each operator, combined with the tuple submission rate, as such information is important to derive the amount of communication over edges. We further extend CEPSim to use LogP model~\cite{culler1993logp} to simulate the network delays between VMs.

Appendix~\ref{app:simulator} validates that our designed simulator can mimic the behavior of the real streaming systems, by empirically comparing the relative performance rank given by our simulator and a real streaming platform.

%\section{Simulator}
%\label{sec:sim}
%\input{simulator.tex}

\section{Experiments}

This section describes our benchmark and baselines (Section~\ref{ssec:setup}), and presents our evaluation results (Section~\ref{ssec:result}).

\subsection{Experimental Settings}
\label{ssec:setup}

\topic{Simulation Environment.} 
We create a cluster in \textit{CEPSim} with $5$ homogeneous devices. The computing capacity of each device is $2.5\mathrm{E}3$ \defn{million instructions per second (MIPS)}. The link bandwidth between devices is $1000$ Mbps. 

\topic{Benchmark Data Set Construction and Statistics.}
We create a new benchmark with $3,150$ graphs in the data set. Graphs vary in the graph topology and number of operators. For example, the graph in Figure~\ref{fig:g_x} illustrates the structure of the graphs with 3 branches and 8 operators in total. Both the number of branches and the longest length of a branch in a graph vary from $2$ to $12$.  Figure~\ref{fig:graph_stats}(a) shows the distribution of the number of operators in the data set ranging from 4 to 26. Within each graph, we randomly assign the CPU utilization and payload to each operator. Note that the CPU utilization of an operator is calculated as $\frac{IPT * R}{MIPS}$, where $IPT$ is the number of instruction per tuple and $R$ is the tuple rate of this operator. Figure~\ref{fig:graph_stats}(b) presents the CPU utilization distribution of all the operators in the 3,150 graphs. As shown in Figure~\ref{fig:graph_stats}(c), we also calculate the data saturation rate at each edge using $\frac{P * R}{BW}$, where $P$ is the payload and $BW$ is the link bandwidth. In practice, the average CPU utilization and payload of an operator can be approximated from a profiling run of the stream processing graph. 
Figure~\ref{fig:graph_stats} demonstrates that our data set provides good coverage of different operator count, CPU utilization, and payloads, which represents a wide range of real-world streaming applications. 
For example, our dataset covers many data-parallel and pipeline-parallel streaming graphs similar to the building blocks of many large real-world stream processing workloads.\footnote{For a graph that is significantly larger than those in our dataset, a common approach~\cite{mirhoseini2017device} is to hierarchically segment it into groups and reduced it to a smaller one covered in our dataset, where each group is treated as a single node.} 
We show that the trained model on the stream graph dataset using our method can capture the graph topologies mostly seen for stream processing. 
                                 
\topic{Baselines.} In the evaluation, we compare our proposed approach with the following baselines:

\noindent $\bullet$ \defn{Encoder-decoder}~\cite{mirhoseini2017device} is an LSTM-based \textit{sequence-to-sequence model} designed for device placements in Tensorflow graphs. This scheduling work for Tensorflow graphs learns one model for each separate graph. When a new graph comes, it needs to repeat the whole time-consuming learning process for the new graph. In comparison, we train our model on many different streaming graphs. Our model learns to capture the meta-information about the graph topologies. We adopt an adapted version of the Tensorflow scheduling work for our task. %this model

\noindent $\bullet$ \defn{METIS}~\cite{karypis1998fast} is a graph partitioning library, which takes the input graph, the computational cost of each operation, the amount of data flowing through each edge, and the number of partitions to produce a mapping of operators to partitions. We set the number of partitions to $5$, same as the number of available devices.

%distributed
\noindent $\bullet$ \defn{IBM Streams}~\cite{streams} is a  streaming platform used in production. We use the \textit{fuser} component from IBM Streams to partition the stream graphs. \textit{Fuser} uses simple rules to balance the number of operators and reduce edge cut among partitions. Same as METIS, it requires the number of partitions as an input parameter, which is set to $5$.  

One advantage of our method is that it can \emph{automatically determine the best number of devices to use}, which makes it more suitable to the varying types of graphs at testing.
To demonstrate this advantage, we also compare our results with the two baselines with their oracles on the best number of devices to use. Specifically, we vary the number of partitions of {METIS} or {IBM Streams} from $2$ to $5$, run their results with each number, compute the throughput, and select the highest throughput to report. The two approaches are denoted as \defn{METIS Oracle} and \defn{IBM Streams Oracle}.

\topic{Hyperparameters.} We randomly select $2,520$ graphs for training and the remaining $630$ graphs for testing. The number of hops $K$ in graph embedding is $2$, and the length of node embeddings is $512$. The network is trained for 40 epochs using Adam optimizer with learning rate $0.001$. At each training step, only one graph is fed to the network. The number of samples $N$ for a training graph varies from $3$ to $6$ (with $3$ on-policy samples and up to $3$ samples from memory buffer). These settings are selected via cross-validation.

\begin{figure}[h]
\includegraphics[trim=0cm 1cm 0cm 1cm, width=\columnwidth]{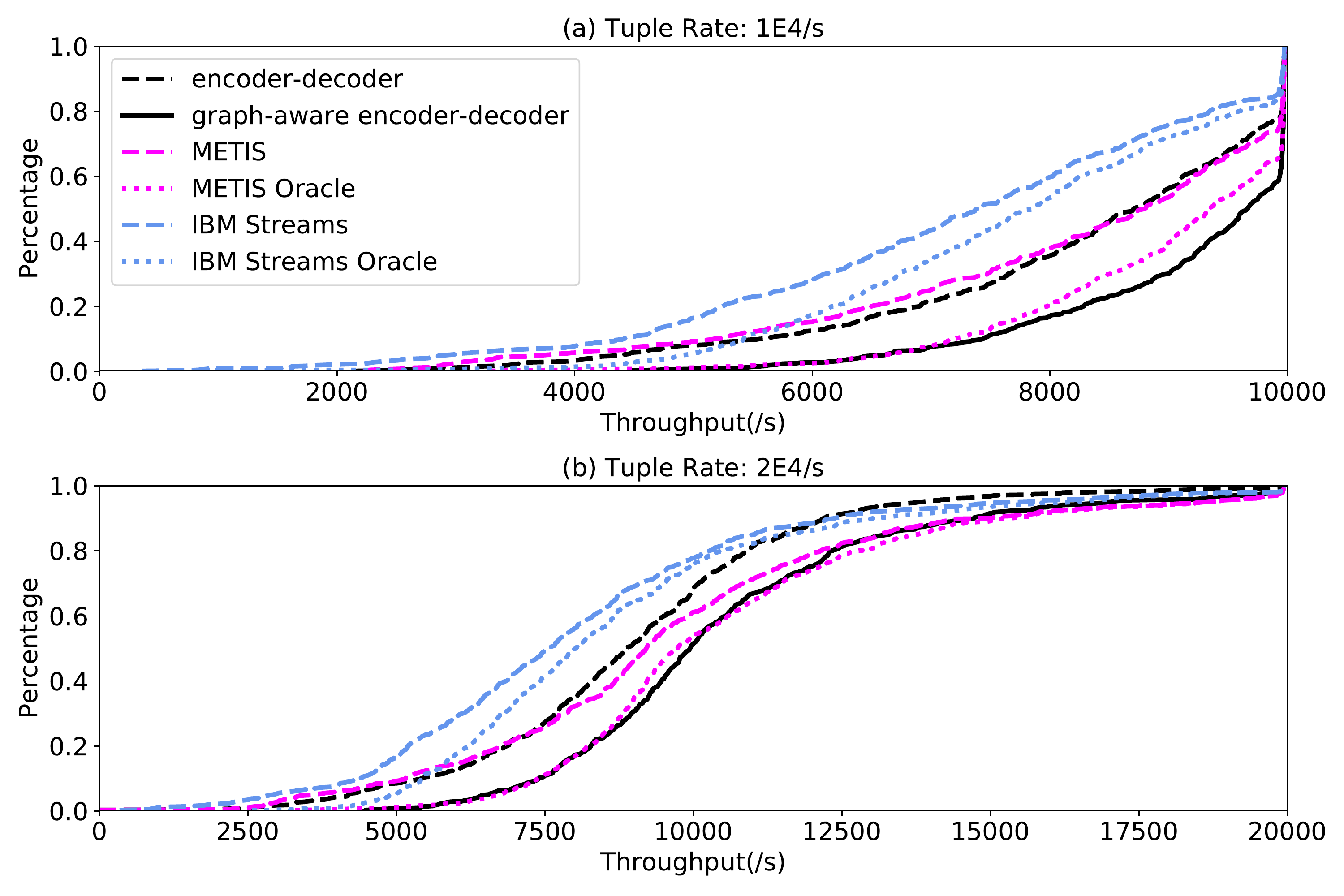}
\caption{CDF of the throughput comparison of our approach with baselines. For instance, ($8000$, $0.17$) means that $17\%$ of graphs have throughput less than $8000/s$. If the CDF plot is more skewed towards right, the overall throughput is better.} 
\label{fig:results}   
\vspace{-6mm}
\end{figure}

\subsection{Results}
\label{ssec:result}

Figure~\ref{fig:results} shows the comparison of the throughput obtained using resource allocations predicted by our graph-aware encoder-decoder model and other baselines over the test data. The source tuple rates of the systems (set in \textit{CEPSim}) are $1\mathrm{E}4/s$ for a typical workload and $2\mathrm{E}4/s$ for stress testing to show that our allocation scheme is reasonable for different tuple rates. In Figure~\ref{fig:results}(a), the minimum throughput of our proposed model is $4468/s$ which is significantly better than other baselines (IBM Streams: $360/s$ METIS: $2220/s$ encoder-decoder: $2140/s$). More than $50\%$ graphs have throughput above $9700/s$ using our approach. In contrast, for the same throughput value, IBM Streams, METIS and encoder-decoder drops to $16\%$, $30\%$ and $27\%$, respectively. Figure~\ref{fig:results}(b) shows similar performance trend between different approaches. This results also confirm the importance of graph-aware encoding and graph-aware decoding in predicting the resource allocation for stream processing graphs. The encoder-decoder model by itself, which is used in tensorflow scheduling work, fails to capture the general graph topologies in our dataset and hence our model outperforms it by a large margin.

\begin{figure}[!h]
\vspace{-2mm}
\includegraphics[trim=0cm 1cm 0cm 0cm, width=\columnwidth]{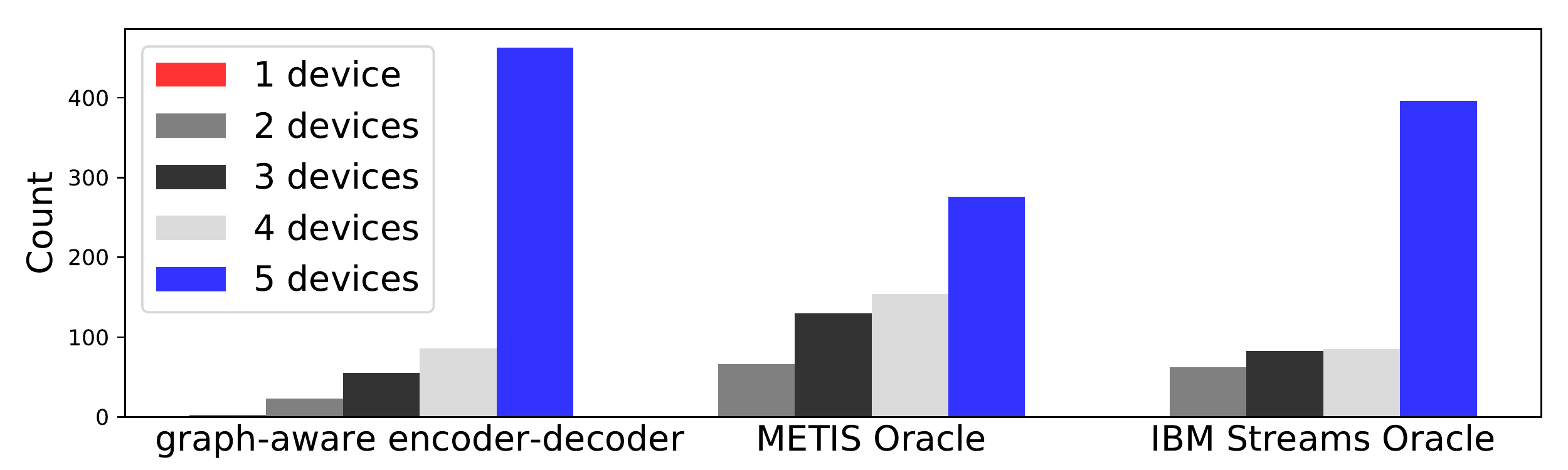}
\caption{Device count distribution of different approaches.} 
\label{fig:results2}   
\vspace{-2mm}
\end{figure}

Moreover, the graph-aware encoder-decoder model exceeds the performance of METIS Oracle and IBM Streams Oracle without human interference to select the optimal partition number. Figure~\ref{fig:results2} presents the device count comparison between these approaches. Unlike METIS and IBM Streams, our model is able to automatically find the right number of devices based on the workload of stream graphs.

\topic{Ablation Study.}
Table~\ref{tab:ablation} shows the ablation study of components in the graph-aware encoder-decoder model. B refers to the baseline LSTM encoder-decoder model. F1 replaces the LSTM encoder in the baseline with graph embedding while decoder remains the same. F2 further enhances the model with the graph-aware decoding. The numbers in the first three columns are relative throughput for source tuple rates $1\mathrm{E}4/s$ ($2\mathrm{E}4/s$). The last column shows the percentage of graphs that have better performance in comparison to METIS\footnote{We choose METIS since it outperforms other baselines except for the artificial oracle ones.}. As can be seen from Table~\ref{tab:ablation}, both the graph encoding and graph-aware decoding improve the performance of resource allocation predictions. Our holistic approach is able to learn better representations of different stream graphs and their complex dependencies. As a result, our model outperforms METIS in $76\%$ ($65\%$) of the test cases.

\begin{table}
\small
\centering
%\vspace{-4mm}
\caption{Ablation study of individual components in our model.}
\vspace{-2mm}
\label{tab:ablation}
\setlength{\tabcolsep}{4pt}
\begin{tabular}{r  c c c c }\toprule[0.15em]
& average & first quartile  & median  &wrt \small{METIS}
%& &  &  & improved graphs
\\\midrule[0.1em]
METIS & 0.81 (0.48) & 0.70 (0.37) & 0.88 (0.46) & - \\\midrule[0.08em]
B & 0.82 (0.45) & 0.73 (0.37) & 0.87  (0.44) & 51\% (39\%) \\%\midrule[0.08em]
B+F1 & 0.87 (0.48) & 0.80 (0.40) & 0.91 (0.47) & 59\% (47\%) \\%\midrule[0.08em]
B+F1\&F2  & \textbf{0.91 (0.53)} & \textbf{0.87 (0.43)} & \textbf{0.97 (0.50)} & \textbf{76\% (65\%)} \\
% + F2 &&&&\\
% \hline       
\bottomrule[0.15em]
\end{tabular}
\vspace{-6mm}
\end{table}

\eat{\begin{table}
\small
\centering
\caption{Tuple Rate: 2e4}
\label{tab:ablation2}
\begin{tabular}{r | c c c c}\toprule[0.15em]
& avg & @25\%  & @50\%  &wrt METIS
%& &  &  & improved graphs
\\\midrule[0.1em]
METIS & 0.48 & 0.37 & 0.46 &  - \\\midrule[0.08em]
encoder-decoder & 0.45 & 0.37 & 0.44 &  39\% \\\midrule[0.08em]
graph-embedding & 0.48 & 0.40 & 0.47 &  47\% \\
decoder & & & & \\\midrule[0.08em]
graph-aware  & \textbf{0.53} & \textbf{0.43} & \textbf{0.50} &  \textbf{65\%} \\
encoder-decoder & & & & \\
\hline       
\end{tabular}
\end{table}}

\topic{Qualitative Study}
Finally, we show examples of different resource allocations from different models, providing details on why the allocation generated by our approach performs better. For each resource allocation scheme, we also give the corresponding throughput number on the top.

\begin{figure}[h]
\centering
\includegraphics[width=\columnwidth]{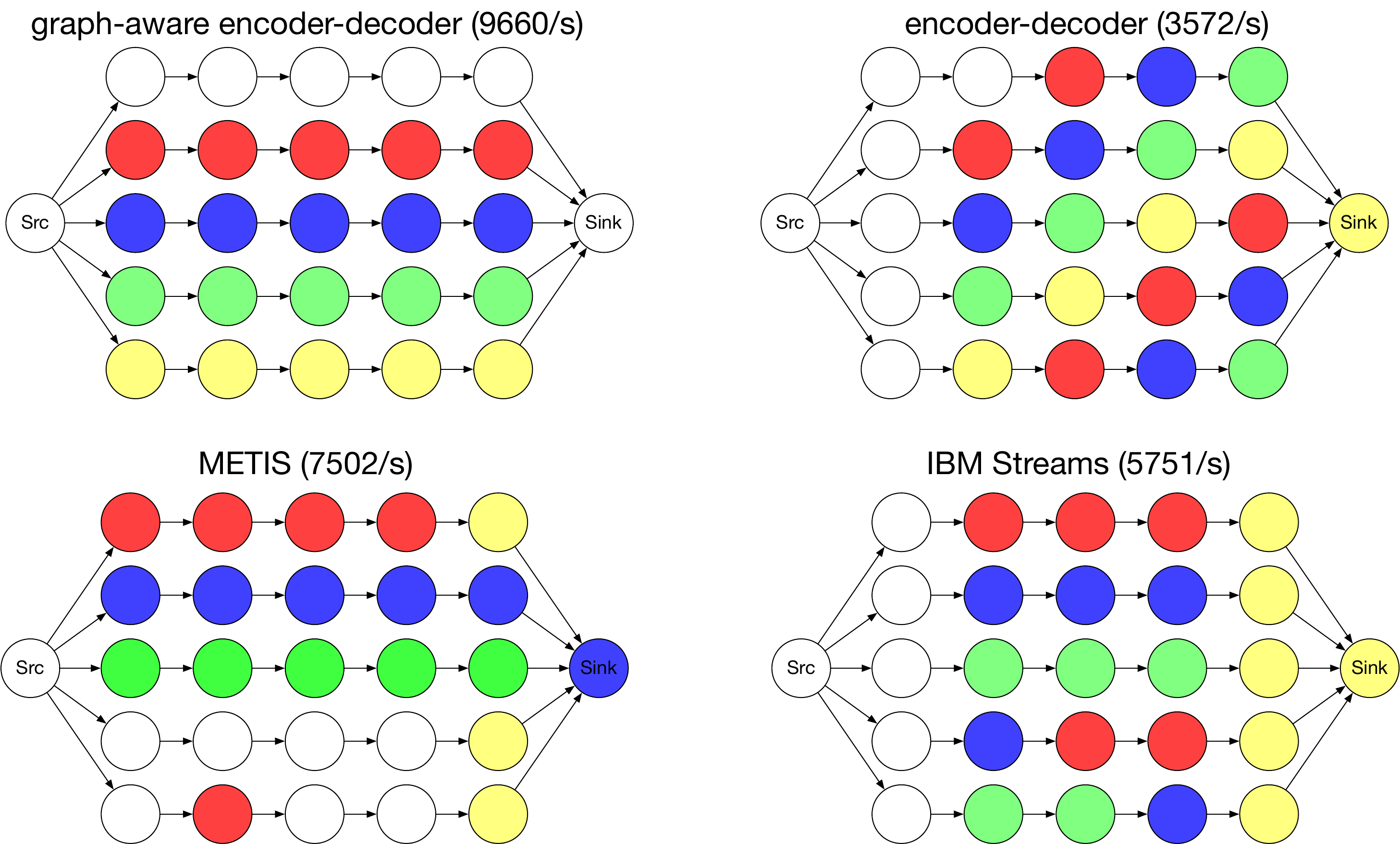}
\vspace{-4mm}
\caption{Example 1. Graph-aware encoder-decoder model outperforms other baselines by colocating operators in the same branch.}
\label{fig:P1}
\vspace{-2mm}
\end{figure}

\begin{figure}[t]
\centering
\includegraphics[width=0.85\columnwidth]{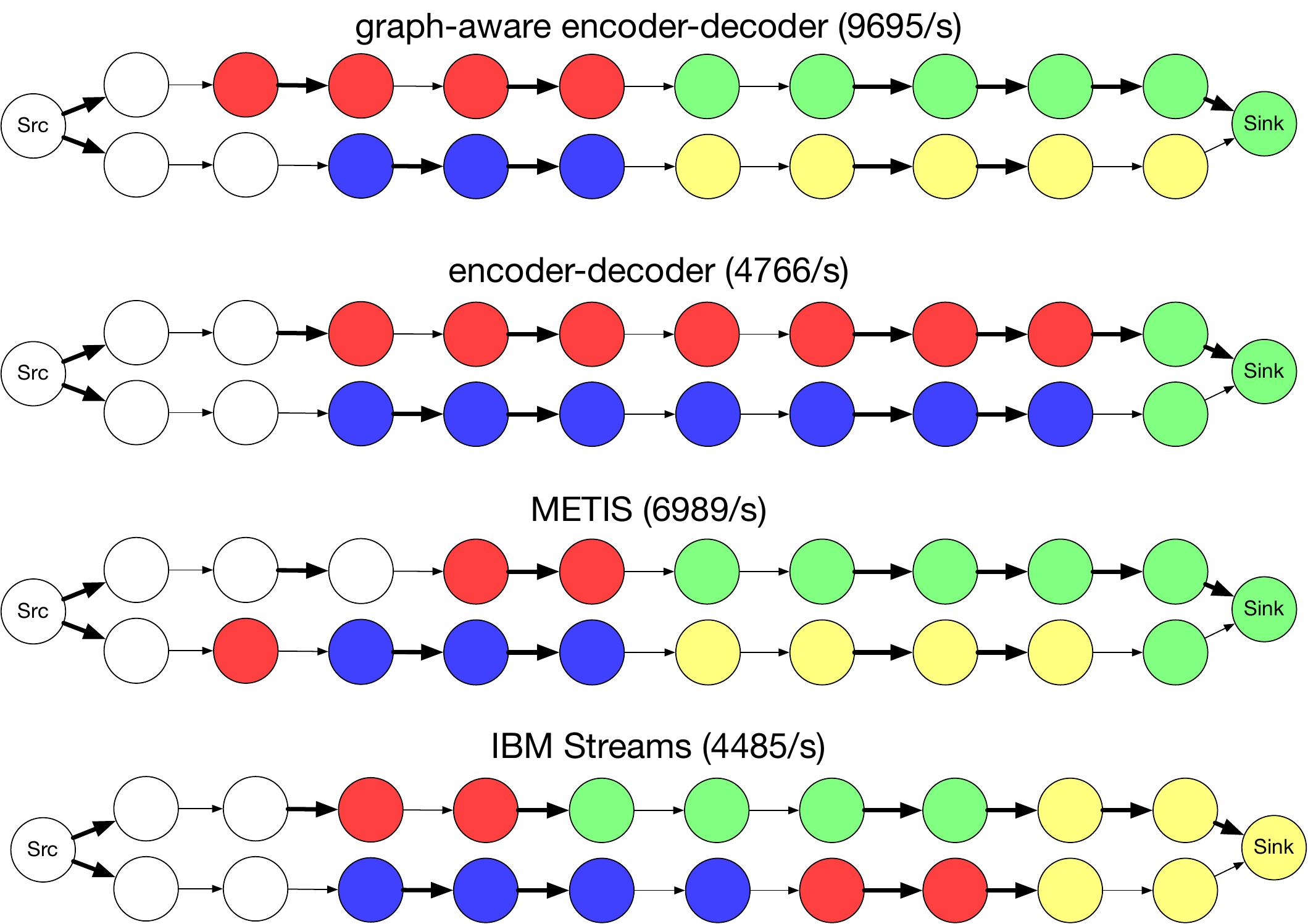}
\vspace{-3mm}
\caption{Example 2. Graph-aware encoder-decoder model outperforms other baselines by avoiding cut the heavy edge while balancing the workload in each partition.}
\label{fig:P3}
\vspace{-4mm}
\end{figure}

Figure~\ref{fig:P1} gives an example where the streaming graph has a relatively larger number of branches, while the communication costs between operators have equal weights. In this situation, an approach that is aware of the global topology is more likely to find the optimal solution, i.e., colocating the operators in the same branch. Our results confirm that our graph encoder is able to capture such topological information, given its perfect prediction. The METIS method, which also benefits from the graph topology information, outperforms the other baselines as well.

Figure~\ref{fig:P3} gives another example with less branches but different weights on the edges. It is found that our method learns to avoid cutting the heavy edges. METIS is also able to avoid cutting heavy edges, however unlike ours, it is not a learning method and does not balance the workload in each partition. More examples can be found in Appendix~\ref{app:analysis}.

\section{Related Work}
\label{sec:related_work}

Here we review and differentiate our approach from the prior work that is most related to this research.

\topic{RL for Job-level Resource Allocation.}
\cite{naik2015performance} proposed an RL-based MapReduce scheduler in heterogeneous environments, which monitors the system state of task execution and suggests speculative re-execution of the slower tasks on other available nodes in the cluster.
\cite{liu2017hierarchical} proposed a hierarchical framework to solve the overall resource allocation and power management problem in cloud computing systems with DRL.
\cite{mao2018learning} introduced a general-purpose scheduling service for data processing jobs in clusters that converts DAGs of jobs to vectors using graph embedding and calculates priority scores for each job stage.

These studies differ from our work significantly in terms of the granularity of resource allocation. They focused on job-level scheduling without dealing with the internal structures of jobs, while we address the resource allocation inside jobs' computation graphs (i.e., operator-level).
As a result, their state representations rely on the hand-designed system- and job-level features, while in this work it is critical to learn state representations from the structured input data.

\topic{RL for Inner-Job Resource Allocation.}
To the best of our knowledge,~\cite{mirhoseini2017device,mirhoseini2018hierarchical,li2018model} are the only works that focus on a similar setting of inner-job allocation like our work.
\cite{mirhoseini2017device,mirhoseini2018hierarchical} use a \emph{sequence-to-sequence model} to predict the device placements for subsets of operations in a TensorFlow graph.
For the device assignment problem for stream processing systems, \cite{li2018model} proposed an actor-critic approach to minimize the average end-to-end tuple processing time.

Our work differs from these work both on the proposed new model and the new problem. 
Most importantly, we focus on a different problem of generalizability to unobserved graphs, while
the foregoing works train and test their models on the same graph. While they consider different applications with different graphs, such as Inception-V3 for image classification, LSTM for language modeling and machine translation and word count application for stream processing, the model for each application is separately trained and tested on its individual graph. 
In comparison, our work aims to find a ``meta'' model that can perform resource allocation on graphs unobserved from training data, which is a more challenging task and is crucial to realistic streaming systems.

\topic{Graph-to-Graph Generation.}
As discussed in Section \ref{sec:background}, our task is a special case of the graph-to-graph generation problem.
Previous works in this direction mainly fall into two types.
First, most works convert the input and output graphs to sequences and reformulate the problem as sequence-to-sequence generation~\cite{iyer2016summarizing,liu2017retrosynthetic}.
The standard sequence models like LSTM-based encoder-encoder can then be directly applied.
Such methods may lose important structural information of the graphs and thus limit their performance.
Second, some works perform direct graph-to-graph generation with techniques like graph VAE~\cite{jin2018junction}.
However, these methods do not handle hard constraints on the topology of the generated graphs.
Since the graphs $G_x$ and $G_y$ in our task have a strong node-level one-to-one dependency, a transformation model considering such a requirement is necessary and is important for improving the performance.
As the above approaches do not suit the properties of our task well, in this work we adopt a new direction based on the graph-to-sequence generation~\cite{xu2018graph2seq} but makes the decoder aware of the generated partial graphs.

There are also works~\cite{khalil2017learning} about general combinatorial search problems on graphs, such as max-cut and TSP. These works do not address the specific encoding/decoding challenges in our task. For example, their policy does not rely on the global graph topology, and their decoder does not need to work with ordered assignments, which does not well suit the real stream applications.

\section{Conclusion}
In this paper, we present a \textit{generalizable} model to predict resource allocation for stream processing systems. We propose to use graph embedding to better represent the structured information of stream graphs, and a graph-aware decoder to capture the complex dependencies that affect the quality of resource allocation. Deep reinforcement learning is applied to jointly train the model with enhanced search efficiency. The proposed framework predicts better resource allocations than graph partitioning library METIS and LSTM-based encoder-decoder model for more than $70\%$ of unobserved graphs, demonstrating superior generalizability of our model. As part of the future work, we plan to apply the proposed framework in the real cloud environment and consider heterogeneous devices in the real deployment.
Besides the decoding formulation, it is possible that the recent graph-to-graph transformation works~\cite{jin2018learning,yu2019dag} can be applied to this task, which we leave for future work.

% \clearpage
\bibliographystyle{aaai}
\bibliography{AAAI-NiX.9089}

% \clearpage

% \renewcommand{\thepage}{\arabic{page}}
% \setcounter{page}{1}

% \documentclass[a4paper]{article}
% \usepackage{appendix}
% \usepackage{times}
% \usepackage{soul}
% \usepackage{url}
% \usepackage[hidelinks]{hyperref}
% \usepackage[utf8]{inputenc}
% \usepackage[small]{caption}
% \usepackage{graphicx}
% \usepackage{amsmath}
% \usepackage{booktabs}

% \begin{document}
% \begin{appendices}
\appendix

\section{Simulator Validation}
\label{app:simulator}

In order to validate that the simulator can mimic the behavior of a real streaming processing system, we compare the relative performance rank given by \textit{CEPSim} and IBM Streams~\cite{streams}, a parallel and distributed streaming platform used in production in dozens of companies in industries, for different graph topologies, operator CPU workloads and tuple payloads.

\begin{figure}[!htbp]
\centering
\includegraphics[width=0.9\columnwidth]{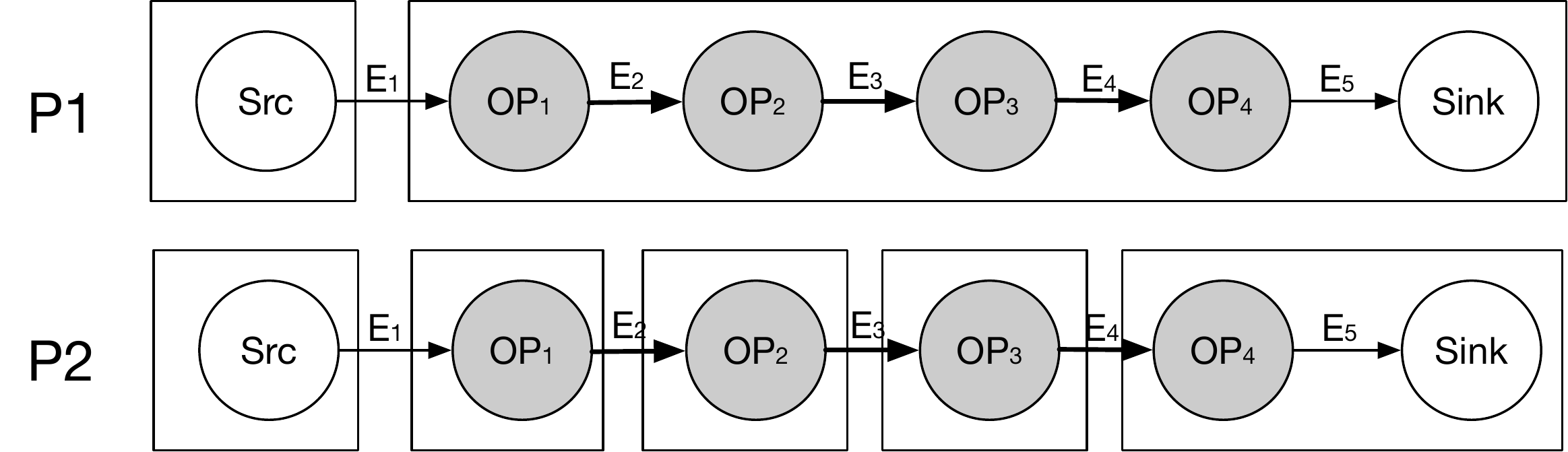}
\vspace{-2mm}
\caption{Resource allocation examples of a pipeline graph.}
\label{fig:graph1}
\vspace{-4mm}
\end{figure}

\begin{table}[!htbp]
\small
\centering
\caption{Rank Comparison for different allocations using pipeline graph. P1 and P2 correspond to the two allocation examples in Figure~\ref{fig:graph1}. The relative rank of different allocations given by CEPSim is the same as IBM Streams. The same trend remains as the operator CPU workload and tuple payload varies.}
\vspace{-2mm}
\label{tab:rank-graph1}
\begin{tabular}{c c c c}\toprule[0.15em]
CPU & Payload & IBM Streams & CEPSim\\\midrule[0.1em]
HIGH & BIG & P1 $<$ P2 & P1 $<$ P2 \\
HIGH & SMALL & P1 $<$ P2 & P1 $<$ P2\\
LOW & BIG & P1 $>$ P2 & P1 $>$ P2\\
LOW & SMALL & P1 $=$ P2 & P1 $=$ P2\\
\bottomrule[0.15em]
\end{tabular}
\end{table}

Figure~\ref{fig:graph1} shows the two types of allocation schemes for a pipeline graph: either we co-locate all the operators except the source operator on the same device (P1) or we try to distribute each operator on five devices (P2). In Table~\ref{tab:rank-graph1}, we vary the CPU workloads of work operators ($OP_1$, $OP_2$, $OP_3$ and $OP_4$) as well as payloads of tuples flowing from those operators. When CPU is set to HIGH in Table~\ref{tab:rank-graph1}, it means in order to catch up with the source tuple rate, the CPU requirement of one work operator is equivalent to the CPU capacity of one device. Vice versa, when CPU is set to LOW, the CPU requirement of one operator is far less than the CPU capacity of one device.
When CPU is HIGH, both Streams and CEPSim makes the right decision to give higher rank for the distributed allocation P2 than the co-location scheme P1. In cases when CPU is LOW, payload plays a more important role determining the good allocation. When tuple payload is set to BIG, P1 performs better than P2 since the outgoing network communication will soon saturate the link bandwidth and hurt the performance. Vice versa when the tuple payload is SMALL, both CEPSim and Streams ranks P1 and P2 the same since the outgoing communication has minimal impact on performance.

\begin{figure}[!htbp]
\centering
\includegraphics[width=\columnwidth]{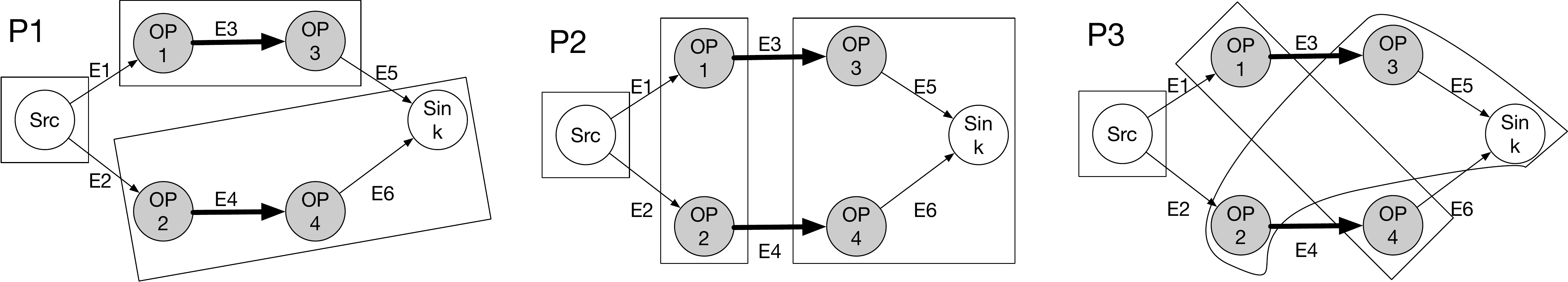}
\vspace{-6mm}
\caption{Resource allocation examples of a graph mixed of data-parallel and pipeline.}
\label{fig:graph2}
%\vspace{-2mm}
\end{figure}

\begin{table}[!htbp]
\small
\centering
\caption{Rank Comparison for different allocations using graph mixed of data-parallel and pipeline. P1, P2 and P3 correspond to the three allocation examples in Figure~\ref{fig:graph2}. The relative rank given by CEPSim for various distributions is the same as Streams.}
\vspace{-2mm}
\label{tab:rank-graph2}
\begin{tabular}{c c c }\toprule[0.15em]
Payload & SIBM treams & CEPSim\\\midrule[0.1em]
BIG & P1 $>$ P3 $>$ P2 & P1 $>$ P3 $>$ P2 \\
SMALL & P1 $=$ P2 $=$ P3 & P1 $=$ P2 $=$ P3 \\
\bottomrule[0.15em]
\end{tabular}
\vspace{-4mm}
\end{table}

Figure~\ref{fig:graph2} shows the three types of resource allocation schemes for a graph mixed with data parallel and pipeline: branch co-location (P1) and branch ex-location (P2, P3). We fix the operator CPU so that it is best to co-locate two work operators to maximize CPU utilization. In Table~\ref{tab:rank-graph2}, we vary the tuple payload flowing from OP1 and OP2, which eventually affects the edge weight of E3 and E4 in Figure~\ref{fig:graph2}. When the tuple payload is SMALL, CEPSim ranks the three allocations the same which is validated by Streams results. When tuple payload is BIG, P1 eliminates network communication, and as a results its performance rank is highest. The total amount communication in P2 and P3 is the same. However, P2 creates communication imbalance while P3 better balance the communication flow between device, as a results the performance rank of P3 is higher than P2 for both CEPSim and Streams.

\section{Analysis on resource allocation prediction}
\label{app:analysis}
Here, we present additional examples of resource allocations computed 
by graph-aware encoder-decoder model, LSTM encoder-decoder model, 
METIS and IBM Streams. For each resource allocation scheme, we also
present the corresponding throughput number.

\begin{figure}[!htbp]
\centering
\includegraphics[width=\columnwidth]{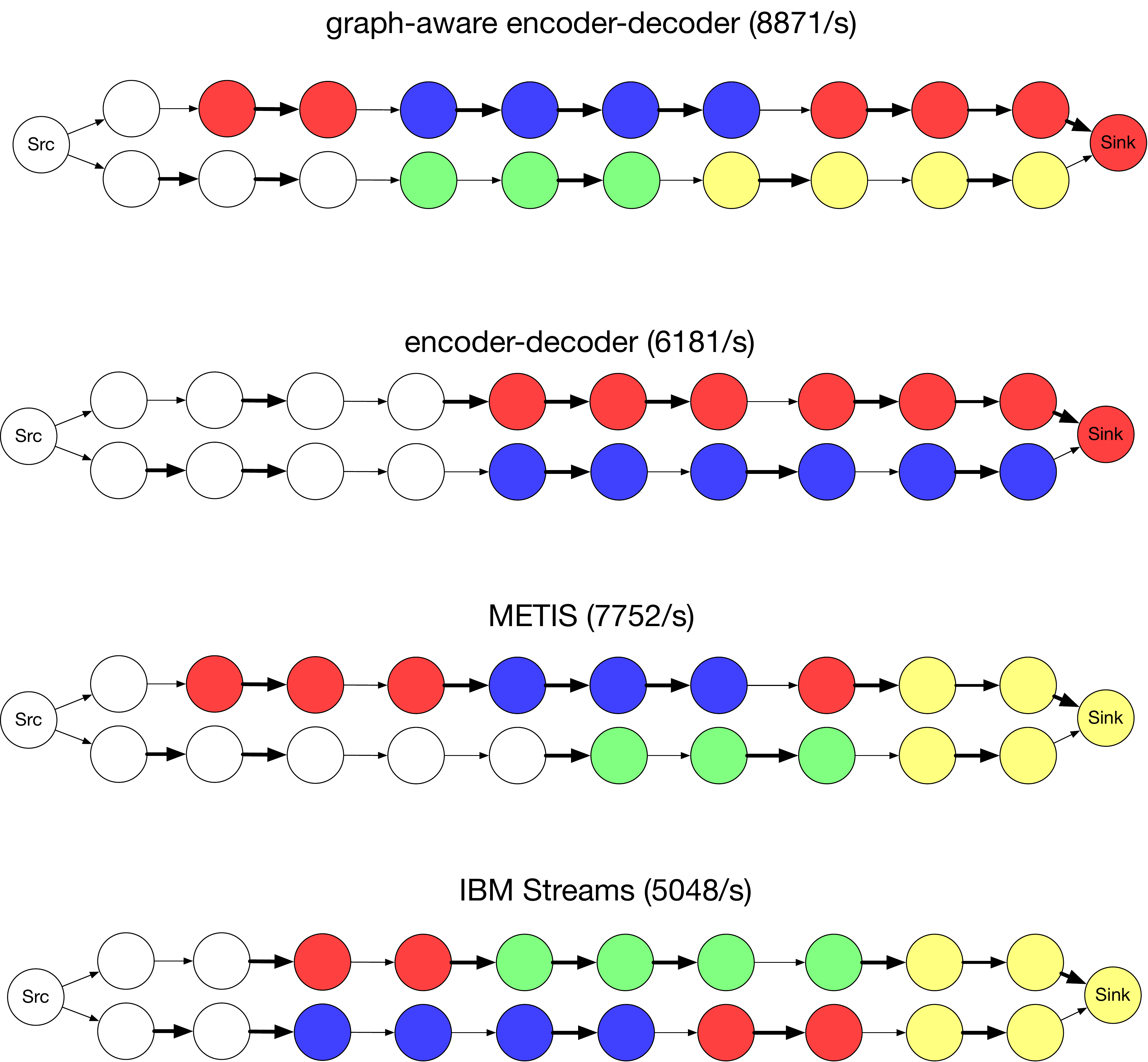}
\caption{Example 3. Graph-aware encoder-decoder model outperforms other baselines by avoiding cut the heavy edge while balancing the workload in each partition.}
\label{fig:P4}
\end{figure}

\begin{figure}[!htbp]
\centering
\includegraphics[width=\columnwidth]{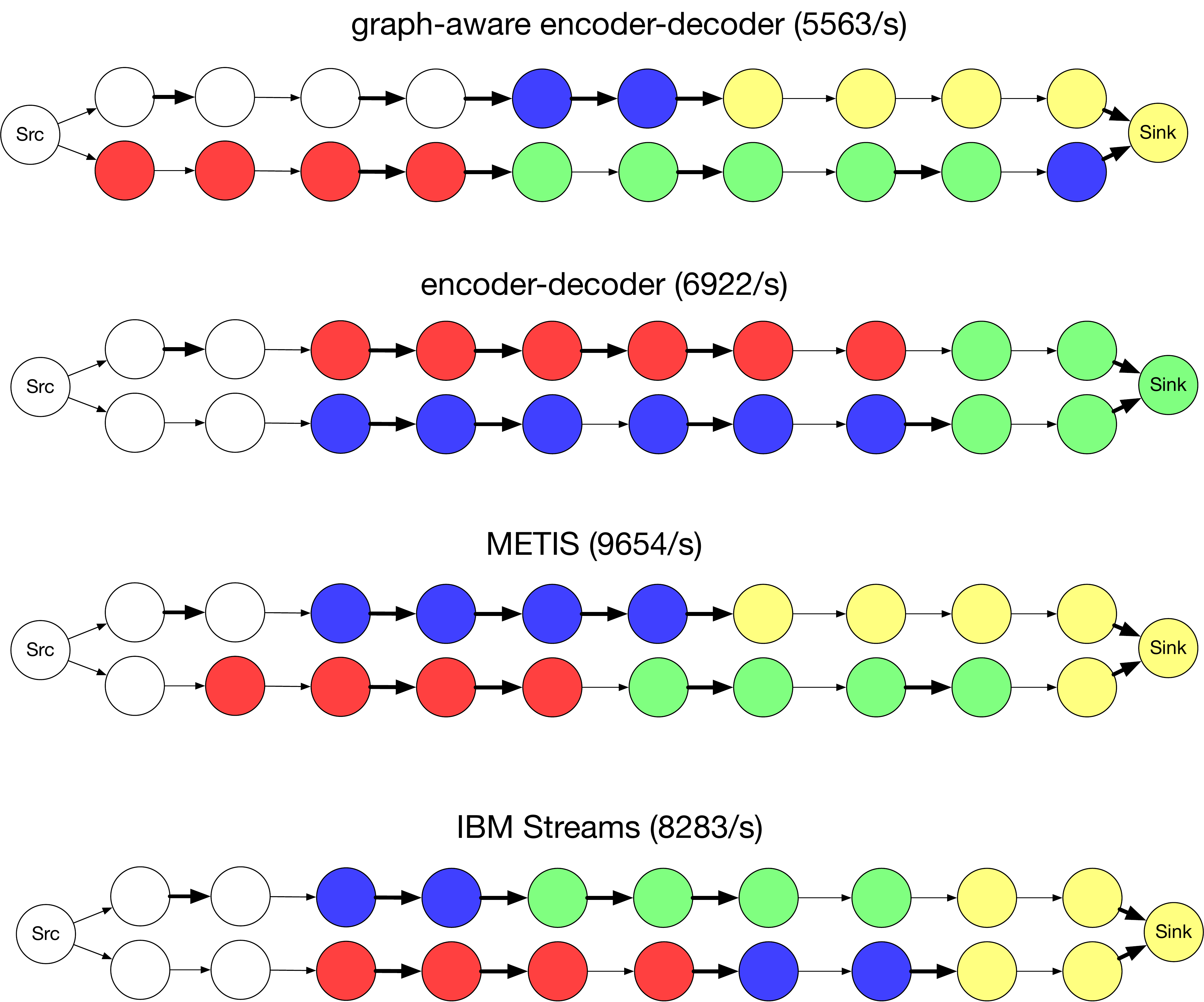}
\caption{Example 4. Graph-aware encoder-decoder model does not perform as well as other baselines since it fails to find partition that avoid cutting heavy edges.}
\label{fig:P5}
\end{figure}

\begin{figure}[!htbp]
\centering
\includegraphics[width=\columnwidth]{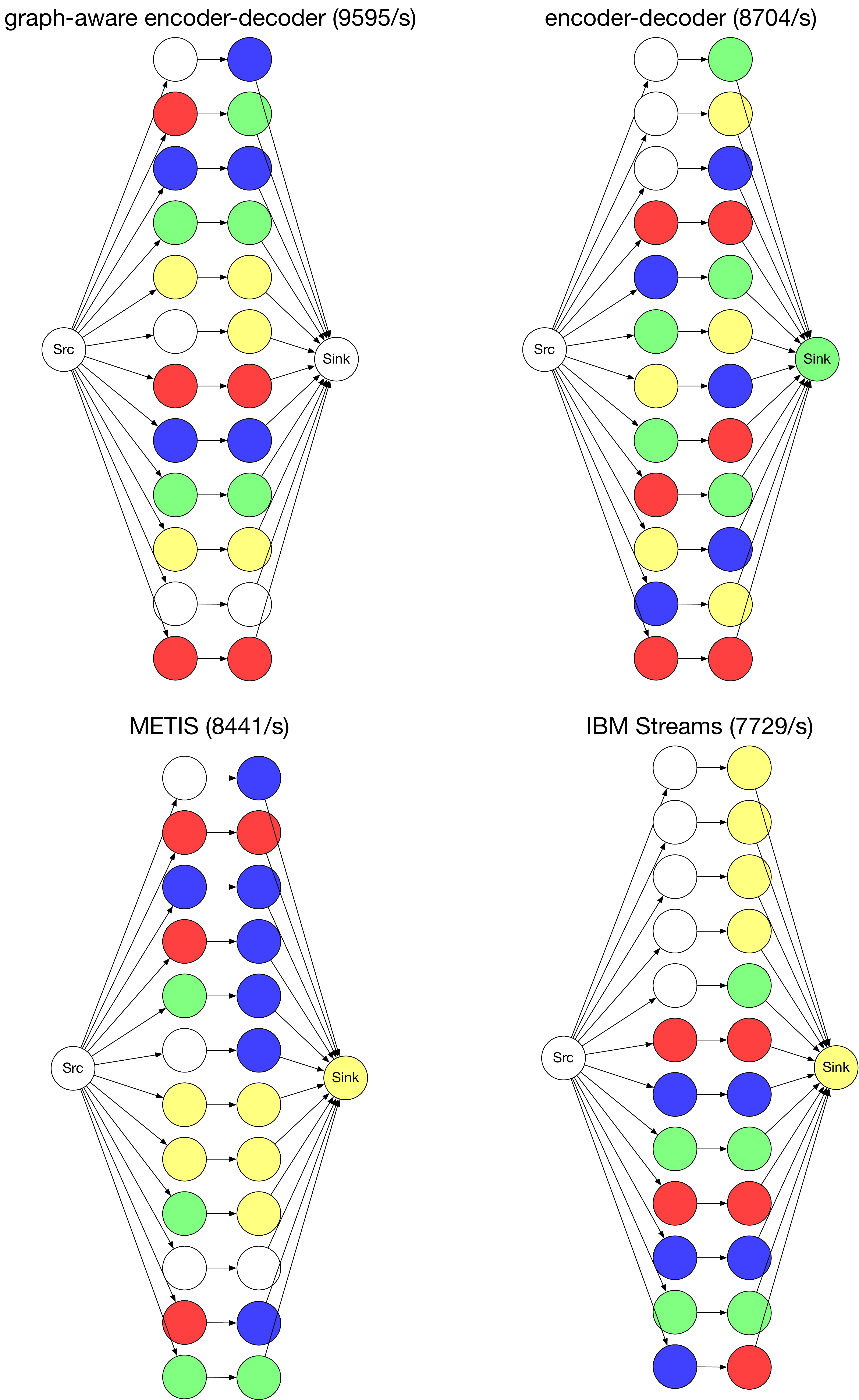}
\caption{Example 5. Graph-aware encoder-decoder model outperforms other baselines by trying to colocate operators in the same branch.}
\label{fig:P2}
\end{figure}

\end{document}